\documentclass[10pt]{article}
\usepackage[latin1]{inputenc}
\usepackage{amsmath}
\usepackage{amsfonts}
\usepackage{amssymb}
\usepackage{mathrsfs}
\usepackage{xcolor}
\usepackage{xcolor}
\usepackage{siunitx}
\usepackage{algorithm}
\usepackage{algorithmic}
\usepackage{amsmath}
\usepackage{rotating} 
\usepackage{amsmath}
\usepackage{array}
\usepackage{booktabs}
\usepackage{booktabs}
\usepackage{multirow}
\usepackage{longtable}
\usepackage{fancyhdr}
\usepackage{blindtext}

\usepackage{MnSymbol}

\usepackage[cal=boondox,scr=boondoxo]{mathalfa}

\usepackage{float}
\usepackage{subfig}
\usepackage{fancybox,graphicx}
\usepackage{subfig}
\usepackage{caption}
\usepackage{color}
\usepackage{authblk}

\usepackage[colorlinks]{hyperref}

\usepackage{hyperref}

\newcommand{\doi}[1]{{doi:~\href{https://doi.org/#1}{\nolinkurl{#1}}}\rmFullStop}

\newcommand*{\rmFullStop}{\rmifnextchar{.}{}{}}

\makeatletter
\newcommand{\rmifnextchar}[3]{%
  \begingroup
  \ltx@LocToksA{\endgroup#2}%
  \ltx@LocToksB{\endgroup#3}%
  \ltx@ifnextchar{#1}{%
    \def\next{\the\ltx@LocToksA}%
    \afterassignment\next
    \let\scratch= %
  }{%
    \the\ltx@LocToksB
  }%
}
\makeatother

\usepackage{accents}
\usepackage[titletoc,title]{appendix}
\usepackage{cite}

\usepackage{mathtools}

\usepackage[top=2in, bottom=1.5in, left=1in, right=1in]{geometry}

\usepackage{stackengine}

\hypersetup{
    linkcolor=black,
}

\title{Adversarial Databases Improve Success in Retrieval-based Large Language Models}

\author[1,2]{Sean Wu}
\author[1,2]{Michael Koo}
\author[2]{Li Yo Kao}
\author[2]{Andy Black}
\author[2]{Lesley Blum}
\author[1,2]{Fabien Scalzo}
\author[2,3,*]{Ira Kurtz}

\affil[1]{Keck Data Science Institute, Pepperdine University, Malibu, 90263 CA, USA.}
\affil[2]{Department of Medicine, Division of Nephrology, David Geffen School of Medicine,
University of California, Los Angeles, Los Angeles 90095-1689 CA, USA.}
\affil[3]{Brain Research Institute, David Geffen School of Medicine,
University of California, Los Angeles, Los Angeles 90095-1689 CA, USA.}

\begin{document}

    \maketitle

\begin{abstract}
Open-source LLMs have shown great potential as fine-tuned chatbots, and demonstrate robust abilities in reasoning and surpass many existing benchmarks. Retrieval-Augmented Generation (RAG) is a technique for improving the performance of LLMs on tasks that the models weren't explicitly trained on, by leveraging external knowledge databases. Numerous studies have demonstrated the effectiveness of RAG to more successfully accomplish downstream tasks when using vector datasets that consist of relevant background information. It has been implicitly assumed by those in the field that if adversarial background information is utilized in this context, that the success of using a RAG-based approach would be nonexistent or even negatively impact the results. To address this assumption,  we tested several open-source LLMs on the ability of RAG to improve their success in answering multiple-choice questions (MCQ) in the medical  subspecialty field of Nephrology. Unlike previous studies, we examined the effect of RAG in utilizing both relevant and adversarial background databases. We set up several open-source LLMs, including Llama 3, Phi-3, Mixtral 8x7b, Zephyr$\beta$, and Gemma 7B Instruct, in a zero-shot RAG pipeline. The source of relevant information was the nephSAP information syllabus from which the MCQ were obtained, and the  UpToDate corpus of clinical information in Nephrology. As adversarial sources of information, text from the Bible and a Random Words generated database were used for comparison. Our data show that most of the open-source LLMs improve their multiple-choice test-taking success as expected when incorporating relevant information vector databases. Surprisingly however, adversarial Bible text significantly improved the success of many LLMs and even random word text improved test taking ability of some of the models. In summary, our results demonstrate for the first time the countertintuitive ability of adversarial information datasets to improve the RAG-based LLM success. The LLM's pre-trained priors are likely involved rather than the RAG mechanism. Whether utilizing adversarial information databases can improve LLM performance in other arenas is a question worthy of pursuing in future research. 
\end{abstract}

\section{Introduction}
LLM chatbots have emerged as one of the most popular applications in the field of Natural Language Processing (NLP)\cite{b1}. Primitive language models were developed from a recurrent neural network (RNN) backbone\cite{b2}; a network architecture to process time series or longitudinal data such as long strings of text by storing data inside the network nodes to influence node processing. Such models were useful for initial next word prediction\cite{b3} and language translation algorithms\cite{b4}. The transformer breakthrough\cite{b5} in 2017 by Vaswani et al. allowed for exponential improvements to be made in almost every NLP task, in which the proposed self-attention\cite{b6} or cross-attention \textbf{ }allowed for larger scaled and more effective language models such as the generative pre-trained transfomer series by OpenAI (GPT)\cite{b7}. With further improvements integrating the transformer architecture over the past five years, some notable LLMs have emerged such as OpenAI's GPT-3.5 Turbo and GPT-4\cite{b9,b8}, Google's PaLM\cite{b10}, Meta's Llama\cite{b11}, and Anthropic's Claude\cite{b12} that have all performed well on various scientific benchmarks including  ScienceQA\cite{b13} and United States Medical Licensing Examination (USMLE)\cite{b14}.  Despite the improving success of proprietary LLMs , open-source LLMs still lag in performance in more specialized fields such as subspecialty medicine test taking ability \cite{b15}. 
\begin{figure}[t!]
  \centering
  \includegraphics[width=1.0\textwidth]{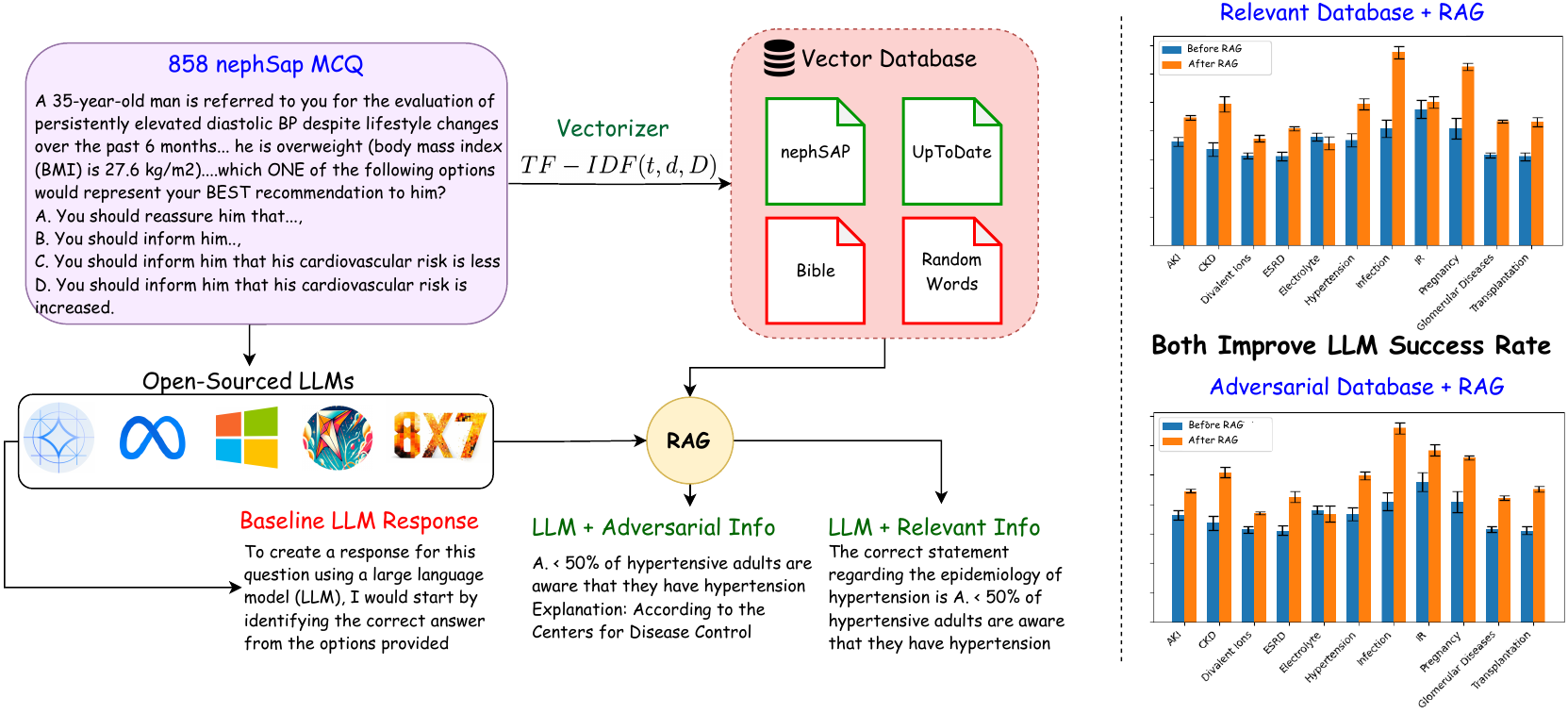}
  \caption{Overall methodology used to demonstrate that in RAG-based settings, adversarial databases counterintuitively can improve the success of correctly answering domain specific MCQ for specific LLMs.}
  \label{fig:example_figure}
\end{figure}
\subsection{Zero-Shot Querying}
A few primitive algorithms that don't require any parameter updates include in-context learning, where $n$ examples are passed in the context window of other examples of similar tasks that were answered to assist the LLM in answering the prompt. Another method is instruction following, which is a similar approach where instead of adding examples in the context window, detailed instructions or system prompts are added beforehand. 

One popular method to improve the performance of LLMs is a zero-shot querying strategy called retrieval-augmented generation (RAG) \cite{lewis2020retrieval}, in which an external knowledge vector database is leveraged to inject relevant context into the LLM at generation time. This mechanism begins with a user prompt or query that  is vectorized by an embedding model. Next, a vector database is systematically searched for the closest distance often using the cosine similarity score in the vector space to the query. The closest vector is retrieved as a context and fed into the model, with the goal of providing relevant information to answer the query correctly. There have been many studies in various domains on how RAG improves the question answering ability of LLMs when given a high-quality vector database to query as for example in legal question answering\cite{wiratunga2024cbr} and financial analysis applications\cite{li2024alphafin}. A recent study also reported the utility of RAG to improve the accuracy of LLMs in medicine \cite{xiong2024benchmarking}. 

\subsection{Problem Definition}
In this paper, we address the unexplored question of how adversarial information databases used by RAG can affect the success of LLMs. To address this question, we re-utilized the dataset from our previous investigation, which consisted of 858 multiple-choice questions and answers in the subspecialty medical field of Nephrology (Nephrology Self-Assessment Program (nephSAP)). Two databases were  created that incorporated relevant Nephrology background information to address the question and answer dataset;  the nephSAP syllabus and the UpToDate Nephrology clinical corpus. Furthermore, two additional  databases were created for comparison that contained adversarial background information with respect to their not \textit{a priori }being expected to improve the LLM test taking ability;  Bible text and a separate Random Words text file. We tested the following open-source LLMs: Mixtral 8x7b, Llama 3, Phi-3, Zephyr$\beta$, and Gemma 7b instruct and compared how the RAG methodology using relevant Nephrology background information compared to adversarial background information in modifying the test-taking  success rate of each LLM. A full pipeline of this research is visualized in Figure 1.

\section{Methods}
In this section, we discuss the data used, its sources, and the criteria that distinguish a corpus as relevant or adversarial for the downstream subspecialty task. We also introduce the various open-source LLMs used in this study, that include Mixtral 8x7b, Gemma 7b Instruct, Llama 3, Zephyr$\beta$, and Phi-3. Additionally, we detail a comprehensive RAG pipeline for all the vector databases used.
\subsection{Databases and Definition of Irrelevance}

We tested the various open-source LLMs test-taking abilities by utilizing 858 multiple-choice questions and answers in the medical subspecialty field of Nephrology (Nephrology self assessment program (nephSAP))\cite{b15}. These patient-oriented questions address various topics in Nephrology. We deployed two relevant databases: nephSAP and UpToDate. The nephSAP syllabus consists of reviews of topics and the latest developments in Nephrology (encompassing information from March 2016 to April 2023). UpToDate is an evidence-based corpus, provides diagnostic and therapeutic information in Nephrology (generated from information available as of March 2023). For non-Nephrology (adversarial) databases, we generated a corpus of Bible text (Latin Vulgate\cite{douay1609}) and in addition a separate random word database using Python's Random Word package. The nephSAP database contains 247,750 lines of text, 1,790,131 words, 60,850 number of unique words, has an average word length of 5.24 characters, and a Flesch-Kincaid grade level of 11.20. UpToDate is a longer corpus with 880,850 lines of text, 7,843,922 words, 47,919 unique words, an average word length of 5.59, and a Flesch-Kincaid grade level of 14.70. The Bible database text  has 118,923 lines, 934,970 words, 17,714 unique words, average word length of 3.97, and a lower reading grade level of 8.70. Finally, the Random Words database text contains 1,912,311 words, with an undefined number of sentences, average word length of 9.55 with a reading level score of 745,823.40 that by definition can be ignored. The  nephSAP and UpToDate have  high attributed reading levels because they are advanced medical corpuses.
\begin{table}[ht]
\centering
\caption{Comparison of databases used for RAG}
\label{tab:combined}
\begin{tabular}{lccccc}
\toprule
\multirow{2}{*}{\textbf{Source}} & \multicolumn{2}{c}{\textbf{Terminology Matching}} & \multicolumn{2}{c}{\textbf{Embedding}} \\
\cmidrule(lr){2-3} \cmidrule(lr){4-5}
 & \textbf{Unique Matches} & \textbf{Overlap (\%)} & \textbf{} & \textbf{GloVe Vector Proximity} \\
\midrule
nephSAP & 903 & 33.3 & \multicolumn{2}{c}{0.80} \\
UpToDate & 733 & 27.1 & \multicolumn{2}{c}{0.80} \\
Bible & 67 & 2.47 & \multicolumn{2}{c}{0.59} \\
Random Words & 240 & 8.86 & \multicolumn{2}{c}{0.06} \\
\bottomrule
\end{tabular}
\end{table}

The nephSAP and UpToDate Nephrology databases were chosen as sources of information that would potentially enable LLMs to more successfully answer the set set of MCQ. The nephSAP database in particular would be predicted to be most informative in this regard because the corpus of information provides a background for the 858 questions. In contrast, the Bible and Random Words databases would not be predicted to provide useful information. 

To quantify how relevant the four text databases are to the field of Nephrology, we first curated a Nephrology term dataset using GPT-4o. This dataset consists of medical terms in Nephrology, totaling 2,709 unique words stored in a set. The first demonstration of relevance involved comparing the number of unique matches in each of the databases. The nephSAP database had the highest number of unique matches (903), followed by UpToDate (733), whereas the Bible had 67 unique matches, and Random Words 240. We also examined the percent overlap of each database, where nephSAP had the highest percentage (33.3\%), followed by UpToDate (27.1\%). Both the Bible and Random Words databases had a minimal overlap (2.47\% and 8.86\%). Relevance and irrelevance can also be quantified in the embedding space. To do this, we deployed a pre-trained GloVe model. GloVe constructs the word vectors of the Nephrology dataset terms and the four databases by factorizing the word co-occurrence matrix. Specifically, we used the glove-wiki-gigaword-50 model from the gensim package to quantify the embeddings and then calculated the cosine similarity scores. The nephSAP and UpToDate both had a score of 0.80, whereas the Bible and Random Words had values of 0.59 and 0.06 respectively. These results are depicted in Table 1.


\subsection{Open-Source Large Language Models}
In this study, we examined several open-source LLMs including  Llama 3, Phi-3, Mixtral 8x7b, Gemma, and Zephyr$\beta$.  Each LLM was deployed either through the HuggingFace pipeline or through Ollama.
\subsubsection{Llama 3}
Llama 3 is a foundation model released in 2024 from Meta AI\cite{meta_Llama3_2024}. It is built off of a standard decoder-only transformer model, and pretrained on over 15 trillion tokens. The Llama 3 model is an improvement from the previous Llama 3 models from Meta AI. In our study, we utilized the 8 billion parameter instruction fine-tuned version of Llama 3 from Huggingface\cite{Llama3modelcard}.
\subsubsection{Phi-3}
We also tested  the smaller open-source LLM Phi-3, a 3.8 billion parameter model from Microsoft's AI team \cite{abdin2024phi}. We utilized the longer context model Phi-3-mini-128k-instruct model from Huggingface (instruction fine-tuned). 
\subsubsection{Mixtral 8x7b}
Mixtral 8x7b  is a ``Mixture of Experts" type of language model that has the same number of parameters as the Minstral 7b model\cite{jiang2024mixtral}. However, each layer in Mixtral 8x7b consists of eight unique feedforward blocks,  each being an ``expert".  We used  Mixtral-8x7B-Instruct-v0.1 from Huggingface.
\subsubsection{Gemma}
Gemma from Deepmind, uses the technology as in the Gemini models to configure Gemma from Google Deepmind \cite{team2024gemma}, we utilized the Ollama library. For faster inference times, Gemma 7b Instruct was loaded with 7 billion parameters and 4 bit quantization.
\subsubsection{Zephyr$\beta$}
Zephyr \cite{tunstall2023zephyr} was trained from distilled supervised fine-tuning (dSFT)\cite{chung2024scaling}. Like Gemma, we utilize OLlama with 4 bit quantization to run the 7 billion parameter model.

\subsection{Retrieval Augmented Generation}
To create a proper retrieval augmented generation (RAG) system, we first converted our text databases into chunks of texts to be later fed as context into the open-source LLMs. Each corpus was divided into  into 1000 word chunks. We then followed a standard pipeline, where for each chunk we obtained the chunk embedding through a TF-IDF vectorizer and stored these into a vector database. In inference time, we used this retrieval mechanism to retrieve the most relevant chunks. Before inputting this context to the LLM, we first formatted the input as ``Context:" ``Question:", then ``Answer:",  similar to an instruction following prompting strategy, and let the open-source LLM respond. For each of the 858 MCQ, we fed the context (patient background information) to the RAG pipeline. We use a TD-IDF vectorizer since it is faster to train and less computationally expensive than BERT\cite{devlin2018bert}. A more formalized representation of our RAG pipeline can be found in the below. For a given open-source LLM, the MCQ success rate was determined for each database four separate times. To compare the effect of RAG using the databases, the means of the four trails and standard errors were analyzed.

\subsubsection{Detailed RAG Workflow}
To visualize the RAG workflow used to test the open-source LLM's abilities with RAG on both relevant and adversarial information, we begin with the following variables as depicted below.

\begin{table}[h!]
    \centering
    \begin{tabular}{ll}
        \toprule
        \textbf{Variable} & \textbf{Description} \\
        \midrule
        $Q$ & Query \\
        $D_i$ & $i$-th chunk of the document corpus, where $i = 1, 2, \ldots, M$ \\
        $\mathbf{v}_Q$ & TF-IDF vector representation of the query $Q$ \\
        $\mathbf{v}_{D_i}$ & TF-IDF vector representation of the chunk $D_i$ \\
        $\text{cosine\_sim}(\mathbf{v}_Q, \mathbf{v}_{D_i})$ & Cosine similarity between the query vector and the chunk vector \\
        \bottomrule
    \end{tabular}
    \label{tab:variables}
\end{table}

\begin{itemize}
    \item The first step of our RAG pipeline was to compute the TF-IDF embedding or vector representation for each patient case for a given MCQ.
    \[
    \mathbf{v}_Q = \text{TF-IDF}(Q)
    \]
    \item Similarly, we computed the TF-IDF vector representation for every chunk of each of the four  information databases (where each chunk consists of approximately 1000 words).
    \[
    \mathbf{v}_{D_i} = \text{TF-IDF}(D_i), \quad \forall i \in \{1, 2, \ldots, M\}
    \]
\end{itemize}

\textbf{Cosine Similarity Calculation}
\begin{itemize}
    \item The next step was to compute the cosine similarity score between the query (patient case information) vector and each vector chunk in the vector space using the following equation.
    \[
    \text{cosine\_sim}(\mathbf{v}_Q, \mathbf{v}_{D_i}) = \frac{\mathbf{v}_Q \cdot \mathbf{v}_{D_i}}{\|\mathbf{v}_Q\| \|\mathbf{v}_{D_i}\|}
    \]
\end{itemize}

\textbf{Retrieval Augmented Methodology}
\begin{itemize}
    \item We used the top 3 chunks (n=3) $\{D_{i_1}, D_{i_2}, D_{i_3}\}$,  chosen based on the highest cosine similarity scores, and the context for augmentation in RAG. In terms of actual implementation, the chunks were  first sorted by cosine similarity scores, then the top 3 were chosen.
    \[
    \{i_1, i_2, i_3\} = \underset{i \in \{1, \ldots, M\}}{\text{arg max top 3}} \left( \text{cosine\_sim}(\mathbf{v}_Q, \mathbf{v}_{D_i}) \right)
    \]
    \item Finally, the three retrieved contexts were concatenated together with newlines to be fed into the RAG workflow.
        \[
    \text{Context} = D_{i_1} + \text{newline} + D_{i_2} + \text{newline} + D_{i_3}
    \]
\end{itemize}

\subsection{Quantifying LLM Outputs}
\begin{figure}[h]
    \makebox[\textwidth][c]{\includegraphics[width=1.0\textwidth]{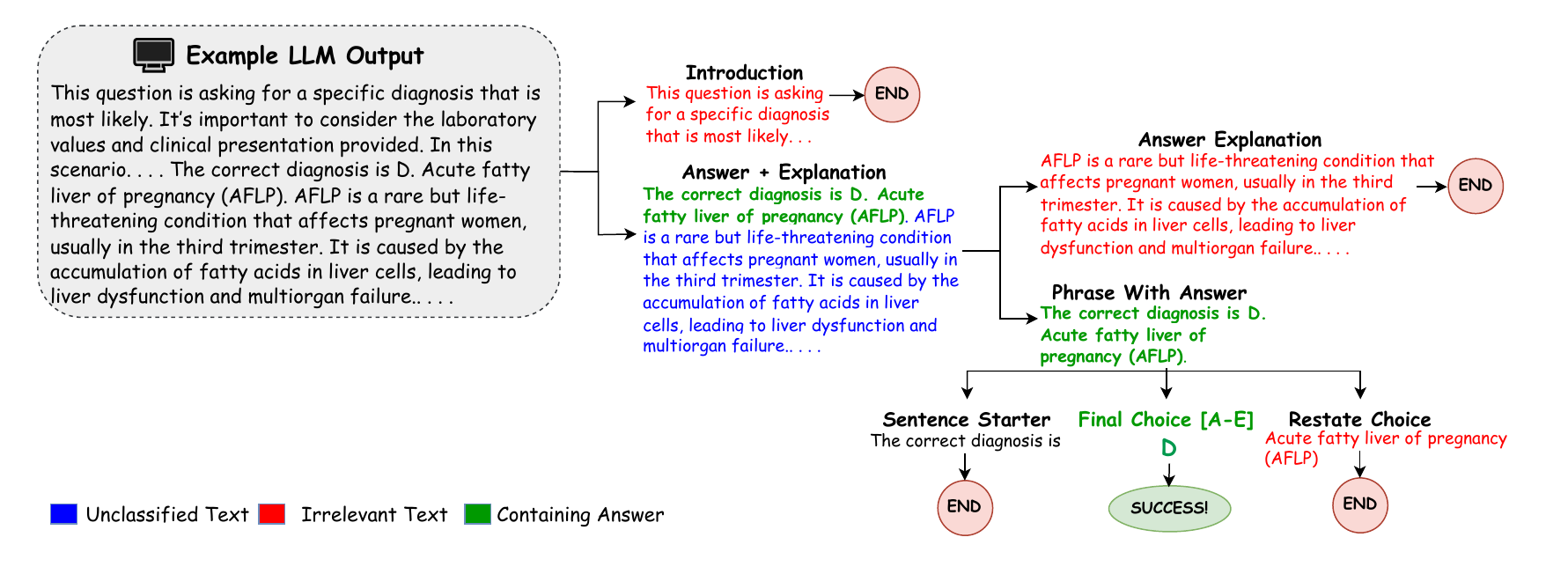} }
    \caption{Example of possible parse tree to automatically extract answer choice from the LLM output. After pattern matching of the introductory phrase and explanatory phrase, the automated script can easily output which answer choice is chosen A-E.}
\end{figure}
To evaluate the performance of the  accuracy of each LLM in answering the questions, we utilized regular expressions to match patterns in the generated outputs and extracted the output answer, and then compared that to the correct answers for each question. Regular expressions enable text processing functions such as validating inputs, extracting data, manipulating strings, and searching/replacing content. We utilized regexes because by utilizing special syntax elements, and complex match patterns we were able to define the patterns we were looking for to compare the outputs with the correct answers. This provided more flexibility than literal text matching alone. 

\subsubsection{Regex Pattern Matching}

We used regular expressions to provide a concise and flexible method, while modifying many different variations of similar patterns for pattern matching to ensure the correct validation of the large dataset of questions being evaluated. However, due to the variability of model text generation in answering questions slightly differently for each question, a large amount of regular expressions were used to ensure accuracy. One example of a regular expression or parse tree to extract the answer from the LLM is shown in Figure 2, where the regex performs pattern matching on the model outputs to correctly detect the answer chosen automatically. By benchmarking against regexes for multiple types of expected patterns, we thoroughly evaluated the different LLM performances.

\section{Results}
In this section we analyze scenarios where adversarial information improved the question answering on some LLMs, did not make a significant improvement, or even made the question answering worse. We also discuss scenarios where the adversarial information effect was not present.
\subsection{Statistics}
For each LLM, the MCQ were answered in four independent experiments. One-way ANOVA and Dunnett's test were used to compare multiple group means. The results are depicted as mean $\pm$ SEM, where $p < 0.05$ was considered significant.  Both overall success and the results for each of 11 Nephrology subcategories (Hypertension, Glomerular Diseases, Acute Kidney Injury (AKI), Divalent Ions, Transplantation, Chronic Kidney Disorder (CKD), End Stage Renal Disease (ESRD), Electrolyte, Pregnancy, Interventional Radiology (IR), and Infection) were analyzed (See Appendix).

As shown in Table 2 and Table 3 using the nephSAP database, most models significantly improved  their test-taking ability except the Phi-3 128k LLM which had no significant change.  The improvement  was model-dependent and varied from  4.1\% (Zephyr$\beta$ 7b)  to 19\% (Mixtral 8x7b). The UpToDate database also improved the test-taking success rate in all LLMs,  0.4\% (Gemma 7b Instruct) to 14.1\% (Mixtral 8x7b) except for the Phi-3 128k LLM where the percent of correctly answered questions actually decreased significantly from 51.4\% to 48.4\%. Finally, there was no clear correlation between the intrinsic  ability of a given LLM and the magnitude of the RAG-based  improvement. 
\begin{table}[t]
\centering
\caption{Mixtral 8x7b RAG percent of MCQ answered correctly}
\label{tab:new_data_results}
\resizebox{\textwidth}{!}{
\begin{tabular}{@{}lccccccc@{}}
\toprule
\multicolumn{8}{c}{\textbf{Mixtral 8x7b}} \\ \midrule
\textbf{Source} & \textbf{Mean (\%)} & \textbf{SEM} & \textbf{vs Baseline} & \textbf{vs Bible} & \textbf{vs nephSAP} & \textbf{vs UpToDate} & \textbf{vs Random} \\ \midrule
Baseline    & 40.2       & 0.34       & --              & $p < 0.001$        & $p < 0.001$          & $p < 0.001$          & $p < 0.001$         \\
nephSAP        & 59.2       & 0.50       & $p < 0.001$     & $p < 0.001$        & --                   & $p < 0.001$          & $p < 0.001$         \\
UpToDate       & 54.3       & 0.22       & $p < 0.001$     & NS                 & $p < 0.001$          & --                   & NS                  \\
Bible          & 54.6       & 0.51      & $p < 0.001$     & --                 & $p < 0.001$          & NS                   & NS                  \\
Random Words   & 55.3       & 0.55      & $p < 0.001$     & NS                 & $p < 0.001$          & NS                   & --                  \\ \bottomrule
\end{tabular}
}
\end{table}
\begin{table}[H]
\centering
\caption{Gemma 7b Instruct RAG percent of MCQ answered correctly}
\label{tab:cosine_scores_gemma}
\resizebox{\textwidth}{!}{
\begin{tabular}{@{}lccccccc@{}}
\toprule
\multicolumn{8}{c}{\textbf{Gemma 7b Instruct}} \\ \midrule
\textbf{Source} & \textbf{Mean (\%)} & \textbf{SEM} & \textbf{vs Baseline} & \textbf{vs Bible} & \textbf{vs nephSAP} & \textbf{vs UpToDate} & \textbf{vs Random} \\ \midrule
Baseline    & 36.8       & 0.27       & --              & $p < 0.05$        & $p < 0.001$          & NS                   & $p < 0.05$         \\
nephSAP       & 41.1       & 0.34       & $p < 0.001$     & $p < 0.001$       & --                   & $p < 0.001$         & $p < 0.001$        \\
UpToDate       & 37.2       & 0.17       & NS              & NS                & $p < 0.001$          & --                   & NS                 \\
Bible          & 38.1       & 0.34       & $p < 0.05$      & --                & $p < 0.001$          & NS                   & NS                 \\
Random Words   & 38.1       & 0.37       & $p < 0.05$      & NS                & $p < 0.001$          & NS                   & --                 \\ \bottomrule
\end{tabular}
}
\end{table}

\begin{table}[h]
\centering
\caption{Zephyr$\beta$ RAG percent of MCQ answered correctly}
\label{tab:cosine_scores_zephyr}
\resizebox{\textwidth}{!}{
\begin{tabular}{@{}lccccccc@{}}
\toprule
\multicolumn{8}{c}{\textbf{Zephyr$\beta$ 7b}} \\ \midrule
\textbf{Source} & \textbf{Mean (\%)} & \textbf{SEM} & \textbf{vs Baseline} & \textbf{vs Bible} & \textbf{vs nephSAP} & \textbf{vs UpToDate} & \textbf{vs Random} \\ \midrule
Baseline    & 29.3       & 0.01       & --              & $p < 0.004$        & $p < 0.001$          & $p < 0.001$         & $p < 0.001$         \\
nephSAP        & 33.4       & 0.004       & $p < 0.001$     & NS                 & --                   & NS                  & $p < 0.001$         \\
UpToDate       & 32.9       & 0.005       & $p < 0.001$     & NS                 & NS                   & --                  & NS                 \\
Bible          & 32.3       & 0.01       & $p < 0.004$     & --                 & NS                   & NS                  & $p < 0.001$         \\
Random Words   & 21.2       & 0.002       & $p < 0.001$     & $p < 0.001$        & $p < 0.001$          & NS                 & --                  \\ \bottomrule
\end{tabular}
}
\end{table}
\begin{table}[h]
\centering
\caption{Llama 3 RAG percent of MCQ answered correctly}
\label{tab:cosine_scores_Llama3}
\resizebox{\textwidth}{!}{
\begin{tabular}{@{}lccccccc@{}}
\toprule
\multicolumn{8}{c}{\textbf{Llama 3 8b}} \\ \midrule
\textbf{Source} & \textbf{Mean (\%)} & \textbf{SEM} & \textbf{vs Baseline} & \textbf{vs Bible} & \textbf{vs nephSAP} & \textbf{vs UpToDate} & \textbf{vs Random} \\ \midrule
Baseline    & 53.7       & 0.17       & --              & $p < 0.002$        & $p < 0.001$          & $p < 0.05$          & $p < 0.001$         \\
nephSAP        & 57.0       & 0.30      & $p < 0.001$     & $p < 0.001$        & --                   & $p < 0.05$          & $p < 0.001$         \\
UpToDate       & 55.4       & 0.42       & $p < 0.05$      & $p < 0.001$        & $p < 0.05$           & --                  & $p < 0.001$         \\
Bible          & 51.7       & 0.33       & $p < 0.002$     & --                 & $p < 0.001$          & $p < 0.001$         & $p < 0.001$         \\
Random Words   & 40.4       & 0.40       & $p < 0.001$     & $p < 0.001$        & $p < 0.001$          & $p < 0.001$         & --                  \\ \bottomrule
\end{tabular}
}
\end{table}
\begin{table}[h]
\centering
\caption{Phi-3 RAG percent of MCQ answered correctly}
\label{tab:cosine_scores_phi3}
\resizebox{\textwidth}{!}{
\begin{tabular}{@{}lccccccc@{}}
\toprule
\multicolumn{8}{c}{\textbf{Phi-3 128k}} \\ \midrule
\textbf{Source} & \textbf{Mean (\%)} & \textbf{SEM} & \textbf{vs Baseline} & \textbf{vs Bible} & \textbf{vs nephSAP} & \textbf{vs UpToDate} & \textbf{vs Random} \\ \midrule
Baseline    & 51.4       & 0.60       & --              & $p < 0.003$        & NS                   & $p < 0.005$         & $p < 0.001$         \\
nephSAP        & 51.0       & 0.23      & NS              & $p < 0.01$         & --                   & $p < 0.05$          & $p < 0.001$         \\
UpToDate       & 48.4       & 0.77      & $p < 0.005$     & NS                 & $p < 0.05$           & --                  & $p < 0.001$         \\
Bible          & 48.2       & 0.50       & $p < 0.003$     & --                 & $p < 0.01$           & NS                  & $p < 0.001$         \\
Random Words   & 42.4       & 0.42       & $p < 0.001$     & $p < 0.001$        & $p < 0.001$          & $p < 0.001$         & --                  \\ \bottomrule
\end{tabular}
}
\end{table}
\subsection{Adversarial Database Effect}
An unexpected and novel finding in our study was the efficacy of RAG using the adversarial Bible and Random Words databases to improve the accuracy of certain LLMs in answering the MCQ accurately.  Specifically, as shown in Table 2, Mixtral 8x7b had  significantly greater overall success using RAG with both the Bible database: 40.2\% versus 54.6\%,  and the Random Words database: 40.2\% versus 55.3\%.  In fact, the efficacy of the Bible database although less than nephSAP was almost identical to UpToDate (Bible: 54.6\% versus UpToDate: 54.3\%). The Random Words database efficacy, 55.3\%, was also less than nephSAP (59.2\%) but similar to the improvement using UpToDate (54.3\%). The approximately 15\% RAG-based significant improvment using irrrelevant information databases with Mixtral 8x7b exceeded all other LLM models.  Similar although lesser improvents with the Bible and Random Words databases were obtained with Gemma 7b Instruct (Table 3), with the Bible database significantly improving the percent of questions answered correctly from 36.8\% to 38.1\% as did the Random Words database from 36.8\% to 38.1\%.  As shown in Table 4, Zephyr Beta 7b significantly improved the RAG-based success rate using the Bible database from 29.3\% to 32.3\% whereas the Random Words database significantly decreased  the percent of correct answers  to 21.2\%.  In both Llama 3 8b (Table 5) and Phi-3 128k (Table 6), both the Bible and Random Words databases significantly decreased the correct answer success rate. These findings suggest that the adversarial information  RAG effect is directionally LLM model dependent.

\section{Discussion}
We have uncovered the novel  phenomenon  that  adversarial information databases are capable of improving  RAG-based LLM  accuracy, and in some instances are essentially equivalent to relevant databases. To our knowledge this effect has not been previously described. Specifically,  we found that adversarial database information was able to significantly improve the the success rate of specific LLMs to answer MCQ accurately in the subspecialty medical field of Nephrology. The finding that various LLMs showed the same phenomenon to various degrees, and that two independent databases (Bible and Random Words) with adversarial information were each effective in certain circumstances, suggest that the phenomenon is not specific to the exact conditions of our experiments. Our findings are potentially generalizable to other domains, where RAG-based approaches are utilized to improve the capability of LLMs. 

\subsection{Role of Attention Mechanism in RAG Effects}
One possible explanation for this phenomenon is the attention mechanism in LLMs. In a non-RAG scenario, the prompt embeddings are passed into a positional encoder, in which each attention score for token $i$ is computed with respect to token $j$ with the query matrix $Q$, key matrix $K$, and value matrix $V$, with also a $d_k$ factor. The following shows how each token attends or weights to all other tokens. 
\begin{figure}[t]
  \centering
  \includegraphics[width=1.0\textwidth]{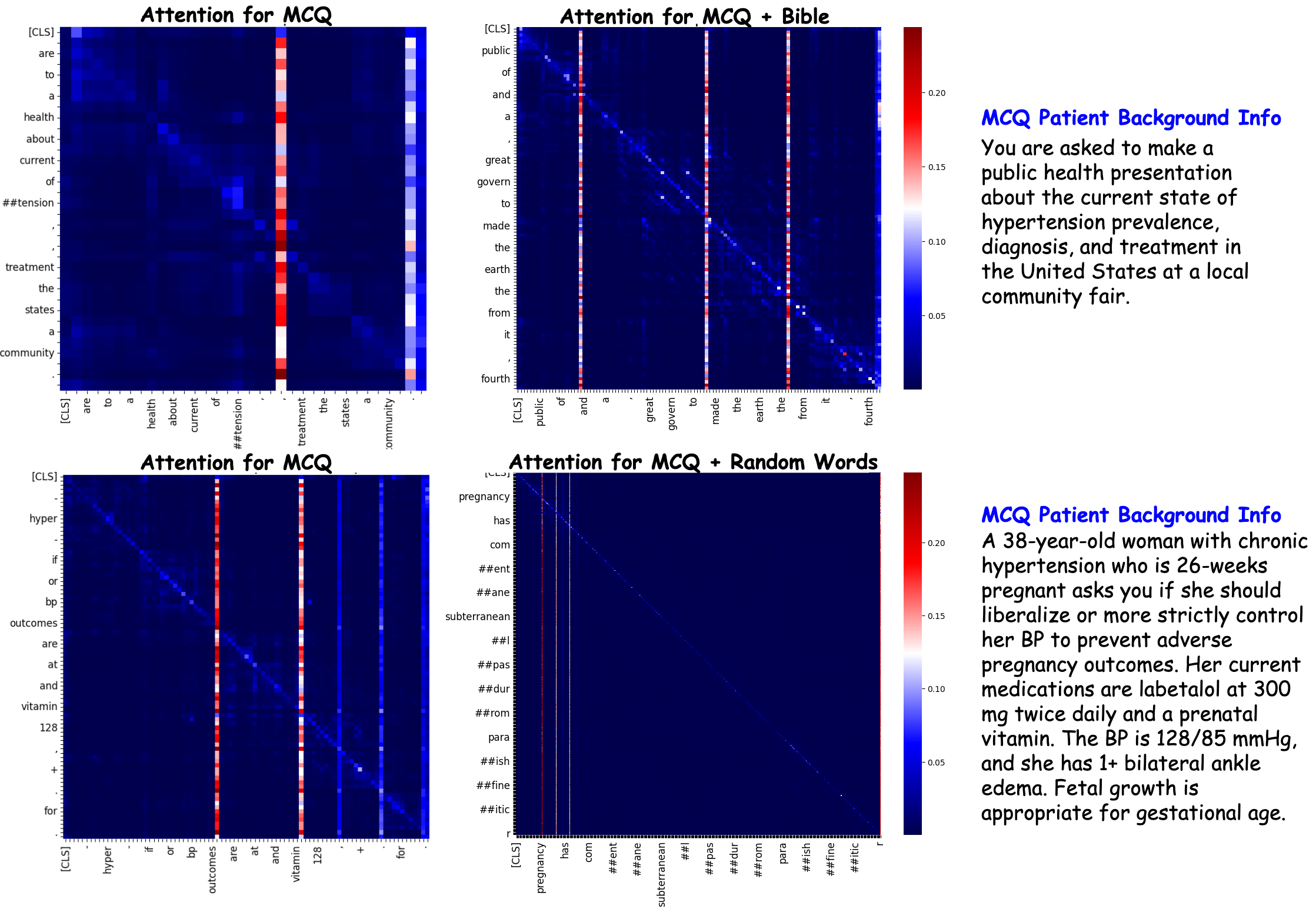}
  \caption{Visualization of DistilBERT attention outputs given both a MCQ prompt and also a Bible or Random Words + MCQ prompt. An evident difference in the weighting matrix is demonstrated. }
  \label{fig:example_figure}
\end{figure}
\begin{equation}
\text{score}_{ij} = \frac{\mathbf{Q}_i \cdot \mathbf{K}_j}{\sqrt{d_k}}
\end{equation}
Accordingly, in a non-RAG scenario where the attention scores of the prompt are computed alone, the attention mechanism is only focused on the prompt. However, when external knowledge is retrieved via RAG, the attention mechanism has more tokens to account for. Therefore, when analyzing the transformer architecture, there is a shift in attention between the tokens in the original prompt even when adversarial information is utilized. When retrieved adversarial information is passed to the multiple-choice question and answer prompt, the latent shift that occurs within the input token representation shifts, which may cause the attention mechanism to emphasize specific parts of the prompt  to a greater degree. These findings are consistent with the results of adversarial attacks on LLMs \cite{sclar2023quantifying}, which demonstrate that even a small change in prompt formatting can produce very different LLM outputs. We demonstrate an example of this attention shift in the mutiple choice question and answer prompt and a section from either the Bible or Random Words databases (Figure 3). To visualize this shift, we deployed a DistilBERT\cite{sanh2019distilbert} model, which is a small scale general pre-trained transformer, and pass both the prompt and the Bible or Random Words + prompt. We then visualized the attention outputs in a heatmap (Figure 3). For simplicity, we extracted the attention outputs from only the final layer. It is evident that, given an excerpt from the Bible database, there is a significant attention shift, which can lead to different and, in this case, improved results. 

\subsection{Implications on RAG and Future Directions}

This work has further implications for future research in retrieval based mechanisms. In some scenarios, the performance of RAG cannot always be attributed to the vector database itself. Importantly, inherent attention mechanisms within the LLM's transformer architecture may come into play. When using RAG, one typically employs databases that contain useful information. Curating these databases can be both time consuming and expensive. Our results suggest that with certain LLMs, it is possible to use non-curated adversarial information to obtain improved LLM results. By simply injecting more tokens into the input stream and shifting the LLM attention span, it might be possible in certain instances improve the accuracy of the the downstream task.

This work has additional implications in the area of retrieval based mechanism research, given that the LLM RAG-based performance cannot always be attributed to the vector database itself. We propose a possible explanation for our finding that is based on LLM attention mechanisms. However, further research is needed to determine whether other underlying mechanisms are involved, so that one can predict in a particular scenario exactly when using an adversarial information database with a RAG-based approach, the success rate of a specific LLM will significantly improve or not.

\section{Conclusion}

In summary,  we tested 858 subspecialty MCQ in Nephrology in various retrieval-based scenarios. We experimented with RAG using databases with relevant background information (nephSAP and UpToDate), as well as adversarial databases (Bible text and Random Words). We found that adversarial databases in certain instances improved the test-taking ability of specific open-source LLMs comparable to relevant information databases.  We highlight the importance of this previously unrecognized novel effect and provide evidence that one potential mechanism involves the injection of more tokens into the input stream as a basis for shifting  LLM attention. All code and data are open-source and available.

\section*{Acknowledgment}
Ira Kurtz is supported in part by funds from the Smidt Family Foundation, Factor Family Foundation, Davita Allen Nissenson Research Fund, Ralph Block Family Fund and Kleeman Family Fund.

\section*{Conflict of Interest declaration}
The authors declare that they have no affiliations with or involvement
in any organization or entity with any financial interest in the subject matter or materials discussed in this
manuscript.

\section*{Author Contributions}
SW contributed to the design, implementation, data analysis and writing the
manuscript; MK contributed to benchmarking the LLMs and writing the manuscript. LK contributed to data analysis. AB and LB contributed to creating the databases. FS contributed to writing the manuscript. IK conceived of the study, and contributed to the
design, implementation, data analysis and writing the manuscript.

\bibliographystyle{ieeetr} 
\bibliography{bibliography.bib}






\appendix
\section*{Appendix}
\section{Compute}
We leveraged Google Colab for cloud CPU and GPU (Nvidia Tesla T4) for preprocessing and running BERT models. Experiments were conducted on a university cluster with eight Nvidia RTX A5000 GPUs, each with 24 GB of memory.
\section{Open-Source Access}
The data and code for this research are available and are open-source. They can be accessed both on our GitHub repository for code on loading the LLMs, benchmarking them, and RAG linked \href{https://github.com/SeanWu25/Adversarial-Databases}{here}, and the vector databases can also be found via Hugging Face in this link \href{https://huggingface.co/datasets/SeanWu25/Adversarial-Databases}{here}.

\section{Analysis on Nephrology Subcategories}
\begin{center}
\small
\begin{longtable}{@{}p{1.3cm}ccccccc@{}}
\caption{Mixtral 8x7b RAG percent of MCQ answered correctly by subcategory}
\label{tab:new_data_results} \\

\toprule
\textbf{Source} & \textbf{Mean (\%)} & \textbf{SEM} & \textbf{vs Baseline} & \textbf{vs nephSAP} & \textbf{vs UpToDate} & \textbf{vs Bible} & \textbf{vs Random} \\
\midrule
\endfirsthead

\multicolumn{8}{c}%
{{\bfseries \tablename\ \thetable{} -- continued from previous page}} \\
\toprule
\textbf{Source} & \textbf{Mean (\%)} & \textbf{SEM} & \textbf{vs Baseline} & \textbf{vs nephSAP} & \textbf{vs UpToDate} & \textbf{vs Bible} & \textbf{vs Random} \\
\midrule
\endhead

\midrule \multicolumn{8}{r}{{Continued on next page}} \\ \midrule
\endfoot

\bottomrule
\endlastfoot

\multicolumn{8}{c}{\textbf{Hypertension}} \\
\midrule
Baseline & 48.9 & 1.62 & -- & $p < 0.001$ & $p < 0.001$ & $p < 0.001$ & $p < 0.001$ \\
nephSAP & 70.8 & 0.79 & $p < 0.001$ & -- & $p < 0.001$ & $p < 0.005$ & $p < 0.01$ \\
UpToDate & 60.4 & 0.28 & $p < 0.001$ & $p < 0.001$ & -- & NS & NS \\
Bible & 62.4 & 2.45 & $p < 0.001$ & $p < 0.005$ & NS & -- & NS \\
Random & 63.2 & 0.84 & $p < 0.001$ & $p < 0.01$ & NS & NS & -- \\
\midrule
\multicolumn{8}{c}{\textbf{Glomerular Diseases}} \\
\midrule
Baseline & 35.2 & 1.24 & -- & $p < 0.001$ & $p < 0.001$ & $p < 0.001$ & $p < 0.001$ \\
nephSAP & 54.9 & 1.46 & $p < 0.001$ & -- & $p < 0.001$ & $p < 0.05$ & NS \\
UpToDate & 46.8 & 1.07 & $p < 0.001$ & $p < 0.001$ & -- & NS & NS \\
Bible & 50.3 & 0.73 & $p < 0.001$ & $p < 0.05$ & NS & -- & NS \\
Random & 50.8 & 0.69 & $p < 0.001$ & NS & NS & NS & -- \\
\midrule
\multicolumn{8}{c}{\textbf{AKI}} \\
\midrule
Baseline & 40.2 & 1.16 & -- & $p < 0.001$ & $p < 0.001$ & $p < 0.001$ & $p < 0.001$ \\
nephSAP & 60.7 & 0.92 & $p < 0.001$ & -- & NS & $p < 0.05$ & $p < 0.001$ \\
UpToDate & 56.2 & 0.46 & $p < 0.001$ & NS & -- & NS & NS \\
Bible & 55.3 & 2.02 & $p < 0.001$ & $p < 0.05$ & NS & -- & NS \\
Random & 51.7 & 1.59 & $p < 0.001$ & $p < 0.001$ & NS & NS & -- \\
\midrule
\multicolumn{8}{c}{\textbf{Divalent Ions}} \\
\midrule
Baseline & 40.5 & 1.67 & -- & $p < 0.001$ & $p < 0.001$ & $p < 0.001$ & $p < 0.001$ \\
nephSAP & 51.5 & 0.98 & $p < 0.001$ & -- & NS & NS & NS \\
UpToDate & 48.5 & 0.98 & $p < 0.001$ & NS & -- & NS & NS \\
Bible & 48.5 & 1.22 & $p < 0.001$ & NS & NS & -- & NS \\
Random & 48.9 & 0.98 & $p < 0.001$ & NS & NS & NS & -- \\
\midrule
\multicolumn{8}{c}{\textbf{Transplant}} \\
\midrule
Baseline & 36.1 & 1.90 & -- & $p < 0.001$ & $p < 0.001$ & $p < 0.001$ & $p < 0.001$ \\
nephSAP & 60.3 & 1.89 & $p < 0.001$ & -- & NS & $p < 0.05$ & NS \\
UpToDate & 54.4 & 2.87 & $p < 0.001$ & NS & -- & NS & NS \\
Bible & 51.7 & 0.72 & $p < 0.001$ & $p < 0.05$ & NS & -- & $p < 0.005$ \\
Random & 56.7 & 1.43 & $p < 0.001$ & NS & NS & $p < 0.005$ & -- \\
\midrule
\multicolumn{8}{c}{\textbf{CKD}} \\
\midrule
Baseline & 45.8 & 1.05 & -- & $p < 0.001$ & $p < 0.001$ & $p < 0.001$ & $p < 0.001$ \\
nephSAP & 63.1 & 2.00 & $p < 0.001$ & -- & NS & NS & NS \\
UpToDate & 58.9 & 0.79 & $p < 0.001$ & NS & -- & NS & NS \\
Bible & 61.1 & 2.36 & $p < 0.001$ & NS & NS & -- & NS \\
Random & 63.3 & 1.20 & $p < 0.001$ & NS & NS & NS & -- \\
\midrule
\multicolumn{8}{c}{\textbf{ESRD}} \\
\midrule
Baseline & 34.2 & 2.77 & -- & $p < 0.001$ & $p < 0.001$ & $p < 0.001$ & $p < 0.001$ \\
nephSAP & 59.4 & 1.16 & $p < 0.001$ & -- & NS & NS & NS \\
UpToDate & 55.3 & 2.28 & $p < 0.001$ & NS & -- & NS & NS \\
Bible & 52.2 & 1.28 & $p < 0.001$ & NS & NS & -- & NS \\
Random & 55.8 & 2.54 & $p < 0.001$ & NS & NS & NS & -- \\
\midrule
\multicolumn{8}{c}{\textbf{Electrolyte}} \\
\midrule
Baseline & 37.9 & 2.99 & -- & $p < 0.001$ & $p < 0.001$ & $p < 0.001$ & $p < 0.05$ \\
nephSAP & 53.0 & 1.63 & $p < 0.001$ & -- & NS & NS & NS \\
UpToDate & 53.0 & 1.91 & $p < 0.001$ & NS & -- & NS & NS \\
Bible & 51.7 & 1.22 & $p < 0.001$ & NS & NS & -- & NS \\
Random & 46.1 & 1.29 & $p < 0.05$ & NS & NS & NS & -- \\
\midrule
\multicolumn{8}{c}{\textbf{Pregnancy}} \\
\midrule
Baseline & 42.5 & 4.17 & -- & $p < 0.05$ & $p < 0.01$ & $p < 0.05$ & $p < 0.05$ \\
nephSAP & 59.2 & 3.44 & $p < 0.05$ & -- & NS & NS & NS \\
UpToDate & 61.7 & 3.97 & $p < 0.01$ & NS & -- & NS & NS \\
Bible & 60.0 & 1.92 & $p < 0.05$ & NS & NS & -- & NS \\
Random & 57.5 & 4.17 & $p < 0.05$ & NS & NS & NS & -- \\
\midrule
\multicolumn{8}{c}{\textbf{IR}} \\
\midrule
Baseline & 46.7 & 3.04 & -- & $p < 0.05$ & $p < 0.01$ & $p < 0.05$ & $p < 0.05$ \\
nephSAP & 60.0 & 2.36 & $p < 0.05$ & -- & NS & NS & NS \\
UpToDate & 62.5 & 4.59 & $p < 0.01$ & NS & -- & NS & NS \\
Bible & 61.7 & 1.67 & $p < 0.05$ & NS & NS & -- & NS \\
Random & 60.0 & 2.36 & $p < 0.05$ & NS & NS & NS & -- \\
\midrule
\multicolumn{8}{c}{\textbf{Infection}} \\
\midrule
Baseline & 45.8 & 6.29 & -- & $p < 0.005$ & $p < 0.05$ & $p < 0.05$ & $p < 0.001$ \\
nephSAP & 67.5 & 2.85 & $p < 0.005$ & -- & NS & NS & NS \\
UpToDate & 60.0 & 3.33 & $p < 0.05$ & NS & -- & NS & NS \\
Bible & 62.5 & 1.60 & $p < 0.05$ & NS & NS & -- & NS \\
Random & 69.2 & 2.10 & $p < 0.001$ & NS & NS & NS & -- \\
\end{longtable}
\end{center}
\begin{center}
\small
\begin{longtable}{@{}p{1.3cm}ccccccc@{}}
\caption{Gemma RAG percent of MCQ answered correctly by subcategory}
\label{tab:gemma_results} \\

\toprule
\textbf{Source} & \textbf{Mean (\%)} & \textbf{SEM} & \textbf{vs Baseline} & \textbf{vs nephSAP} & \textbf{vs UpToDate} & \textbf{vs Bible} & \textbf{vs Random} \\
\midrule
\endfirsthead

\multicolumn{8}{c}%
{{\bfseries \tablename\ \thetable{} -- continued from previous page}} \\
\toprule
\textbf{Source} & \textbf{Mean (\%)} & \textbf{SEM} & \textbf{vs Baseline} & \textbf{vs nephSAP} & \textbf{vs UpToDate} & \textbf{vs Bible} & \textbf{vs Random} \\
\midrule
\endhead

\midrule \multicolumn{8}{r}{{Continued on next page}} \\ \midrule
\endfoot

\bottomrule
\endlastfoot

\multicolumn{8}{c}{\textbf{Hypertension}} \\
\midrule
Baseline & 38.49 & 1.68 & -- & $p < 0.05$ & NS & NS & NS \\
nephSAP & 44.38 & 0.97 & $p < 0.05$ & -- & $p < 0.002$ & NS & $p < 0.05$ \\
UpToDate & 35.40 & 1.41 & NS & $p < 0.002$ & -- & NS & NS \\
Bible & 39.61 & 1.16 & NS & NS & NS & -- & NS \\
Random & 37.64 & 1.86 & NS & $p < 0.05$ & NS & NS & -- \\
\midrule
\multicolumn{8}{c}{\textbf{Glomerular Diseases}} \\
\midrule
Baseline & 29.20 & 0.64 & -- & $p < 0.001$ & $p < 0.001$ & $p < 0.002$ & NS \\
nephSAP & 35.40 & 0.42 & $p < 0.001$ & -- & $p < 0.05$ & $p < 0.001$ & $p < 0.001$ \\
UpToDate & 33.06 & 0.42 & $p < 0.001$ & $p < 0.05$ & -- & NS & $p < 0.003$ \\
Bible & 32.21 & 0.48 & $p < 0.002$ & $p < 0.001$ & NS & -- & $p < 0.05$ \\
Random & 30.20 & 0.47 & NS & $p < 0.001$ & $p < 0.003$ & $p < 0.05$ & -- \\
\midrule
\multicolumn{8}{c}{\textbf{AKI}} \\
\midrule
Baseline & 43.26 & 1.75 & -- & NS & NS & NS & NS \\
nephSAP & 45.79 & 0.96 & NS & -- & NS & NS & NS \\
UpToDate & 45.51 & 0.97 & NS & NS & -- & NS & NS \\
Bible & 43.54 & 0.28 & NS & NS & NS & -- & NS \\
Random & 45.51 & 0.33 & NS & NS & NS & NS & -- \\
\midrule
\multicolumn{8}{c}{\textbf{Divalent Ions}} \\
\midrule
Baseline & 37.39 & 1.67 & -- & NS & NS & NS & $p < 0.05$ \\
nephSAP & 39.60 & 0.98 & NS & -- & $p < 0.003$ & NS & NS \\
UpToDate & 33.18 & 0.85 & NS & $p < 0.003$ & -- & NS & $p < 0.001$ \\
Bible & 37.39 & 1.27 & NS & NS & NS & -- & $p < 0.05$ \\
Random & 42.04 & 0.26 & $p < 0.05$ & NS & $p < 0.001$ & $p < 0.05$ & -- \\
\midrule
\multicolumn{8}{c}{\textbf{Transplant}} \\
\midrule
Baseline & 34.72 & 0.28 & -- & $p < 0.05$ & NS & NS & NS \\
nephSAP & 39.17 & 1.14 & $p < 0.05$ & -- & $p < 0.001$ & $p < 0.001$ & $p < 0.001$ \\
UpToDate & 30.83 & 0.53 & NS & $p < 0.001$ & -- & NS & NS \\
Bible & 32.22 & 0.00 & NS & $p < 0.001$ & NS & -- & NS \\
Random & 31.11 & 1.98 & NS & $p < 0.001$ & NS & NS & -- \\
\midrule
\multicolumn{8}{c}{\textbf{CKD}} \\
\midrule
Baseline & 42.78 & 1.40 & -- & $p < 0.05$ & NS & NS & NS \\
nephSAP & 49.44 & 2.15 & $p < 0.05$ & -- & $p < 0.01$ & NS & $p < 0.002$ \\
UpToDate & 41.67 & 1.06 & NS & $p < 0.01$ & -- & NS & NS \\
Bible & 43.89 & 1.47 & NS & NS & NS & -- & NS \\
Random & 40.56 & 0.72 & NS & $p < 0.002$ & NS & NS & -- \\
\midrule
\multicolumn{8}{c}{\textbf{ESRD}} \\
\midrule
Baseline & 41.11 & 1.20 & -- & NS & $p < 0.05$ & NS & NS \\
nephSAP & 40.56 & 0.72 & NS & -- & $p < 0.05$ & NS & NS \\
UpToDate & 45.28 & 0.53 & $p < 0.05$ & $p < 0.05$ & -- & NS & NS \\
Bible & 44.44 & 0.46 & NS & NS & NS & -- & NS \\
Random & 43.06 & 1.84 & NS & NS & NS & NS & -- \\
\midrule
\multicolumn{8}{c}{\textbf{Electrolyte}} \\
\midrule
Baseline & 32.33 & 0.83 & -- & NS & NS & NS & NS \\
nephSAP & 33.62 & 1.11 & NS & -- & NS & NS & NS \\
UpToDate & 33.19 & 0.43 & NS & NS & -- & NS & NS \\
Bible & 33.19 & 1.09 & NS & NS & NS & -- & NS \\
Random & 32.33 & 0.83 & NS & NS & NS & NS & -- \\
\midrule
\multicolumn{8}{c}{\textbf{Pregnancy}} \\
\midrule
Baseline & 42.50 & 1.60 & -- & NS & NS & NS & $p < 0.01$ \\
nephSAP & 46.67 & 0.00 & NS & -- & $p < 0.003$ & $p < 0.05$ & NS \\
UpToDate & 39.17 & 0.83 & NS & $p < 0.003$ & -- & NS & $p < 0.001$ \\
Bible & 40.83 & 1.59 & NS & $p < 0.05$ & NS & -- & $p < 0.001$ \\
Random & 49.17 & 1.59 & $p < 0.01$ & NS & $p < 0.001$ & $p < 0.001$ & -- \\
\midrule
\multicolumn{8}{c}{\textbf{IR}} \\
\midrule
Baseline & 22.50 & 1.60 & -- & $p < 0.01$ & $p < 0.003$ & $p < 0.01$ & $p < 0.001$ \\
nephSAP & 30.00 & 0.00 & $p < 0.01$ & -- & NS & NS & NS \\
UpToDate & 30.83 & 1.59 & $p < 0.003$ & NS & -- & NS & NS \\
Bible & 30.00 & 1.36 & $p < 0.01$ & NS & NS & -- & NS \\
Random & 31.67 & 1.67 & $p < 0.001$ & NS & NS & NS & -- \\
\midrule
\multicolumn{8}{c}{\textbf{Infection}} \\
\midrule
Baseline & 40.00 & 0.00 & -- & $p < 0.001$ & $p < 0.05$ & $p < 0.05$ & $p < 0.001$ \\
nephSAP & 54.17 & 0.83 & $p < 0.001$ & -- & $p < 0.001$ & $p < 0.001$ & $p < 0.005$ \\
UpToDate & 45.84 & 0.84 & $p < 0.05$ & $p < 0.001$ & -- & NS & NS \\
Bible & 45.00 & 2.15 & $p < 0.05$ & $p < 0.001$ & NS & -- & NS \\
Random & 47.50 & 0.83 & $p < 0.001$ & $p < 0.005$ & NS & NS & -- \\
\end{longtable}
\end{center}
\begin{center}
\small
\begin{longtable}{@{}p{1.3cm}ccccccc@{}}
\caption{Zephyr$\beta$ RAG percent of MCQ answered correctly by subcategory}
\label{tab:zephyr_results} \\

\toprule
\textbf{Source} & \textbf{Mean (\%)} & \textbf{SEM} & \textbf{vs Baseline} & \textbf{vs nephSAP} & \textbf{vs UpToDate} & \textbf{vs Bible} & \textbf{vs Random} \\
\midrule
\endfirsthead

\multicolumn{8}{c}%
{{\bfseries \tablename\ \thetable{} -- continued from previous page}} \\
\toprule
\textbf{Source} & \textbf{Mean (\%)} & \textbf{SEM} & \textbf{vs Baseline} & \textbf{vs nephSAP} & \textbf{vs UpToDate} & \textbf{vs Bible} & \textbf{vs Random} \\
\midrule
\endhead

\midrule \multicolumn{8}{r}{{Continued on next page}} \\ \midrule
\endfoot

\bottomrule
\endlastfoot

\multicolumn{8}{c}{\textbf{Hypertension}} \\
\midrule
Baseline & 31.46 & 0.65 & -- & NS & NS & NS & $p < 0.001$ \\
nephSAP & 33.15 & 2.92 & NS & -- & NS & NS & $p < 0.001$ \\
UpToDate & 35.11 & 2.12 & NS & NS & -- & NS & $p < 0.001$ \\
Bible & 30.90 & 0.73 & NS & NS & NS & -- & $p < 0.001$ \\
Random & 21.35 & 0.46 & $p < 0.001$ & $p < 0.001$ & $p < 0.001$ & $p < 0.001$ & -- \\
\midrule
\multicolumn{8}{c}{\textbf{Glomerular Diseases}} \\
\midrule
Baseline & 25.95 & 0.22 & -- & $p < 0.05$ & $p < 0.05$ & $p < 0.05$ & $p < 0.01$ \\
nephSAP & 31.71 & 1.46 & $p < 0.05$ & -- & NS & NS & $p < 0.001$ \\
UpToDate & 31.55 & 1.13 & $p < 0.05$ & NS & -- & NS & $p < 0.001$ \\
Bible & 31.21 & 1.87 & $p < 0.05$ & NS & NS & -- & $p < 0.001$ \\
Random & 18.62 & 0.57 & $p < 0.01$ & $p < 0.001$ & $p < 0.001$ & $p < 0.001$ & -- \\
\midrule
\multicolumn{8}{c}{\textbf{AKI}} \\
\midrule
Baseline & 35.21 & 0.75 & -- & NS & NS & NS & $p < 0.001$ \\
nephSAP & 35.12 & 0.54 & NS & -- & NS & NS & $p < 0.001$ \\
UpToDate & 33.43 & 0.71 & NS & NS & -- & NS & $p < 0.001$ \\
Bible & 35.11 & 0.28 & NS & NS & NS & -- & $p < 0.001$ \\
Random & 25.00 & 1.25 & $p < 0.001$ & $p < 0.001$ & $p < 0.001$ & $p < 0.001$ & -- \\
\midrule
\multicolumn{8}{c}{\textbf{Divalent Ions}} \\
\midrule
Baseline & 22.42 & 2.36 & -- & NS & NS & $p < 0.05$ & $p < 0.05$ \\
nephSAP & 23.89 & 1.40 & NS & -- & NS & NS & $p < 0.01$ \\
UpToDate & 24.12 & 0.84 & NS & NS & -- & NS & $p < 0.004$ \\
Bible & 28.10 & 0.56 & $p < 0.05$ & NS & NS & -- & $p < 0.001$ \\
Random & 17.04 & 1.22 & $p < 0.05$ & $p < 0.01$ & $p < 0.004$ & $p < 0.001$ & -- \\
\midrule
\multicolumn{8}{c}{\textbf{Transplant}} \\
\midrule
Baseline & 25.55 & 1.70 & -- & $p < 0.001$ & $p < 0.05$ & $p < 0.001$ & $p < 0.01$ \\
nephSAP & 34.44 & 0.79 & $p < 0.001$ & -- & NS & NS & $p < 0.001$ \\
UpToDate & 31.11 & 0.79 & $p < 0.05$ & NS & -- & NS & $p < 0.001$ \\
Bible & 33.33 & 0.45 & $p < 0.001$ & NS & NS & -- & $p < 0.001$ \\
Random & 19.45 & 1.47 & $p < 0.01$ & $p < 0.001$ & $p < 0.001$ & $p < 0.001$ & -- \\
\midrule
\multicolumn{8}{c}{\textbf{CKD}} \\
\midrule
Baseline & 30.00 & 2.31 & -- & $p < 0.001$ & $p < 0.001$ & $p < 0.05$ & NS \\
nephSAP & 39.17 & 0.53 & $p < 0.001$ & -- & NS & NS & $p < 0.001$ \\
UpToDate & 40.00 & 0.79 & $p < 0.001$ & NS & -- & $p < 0.05$ & $p < 0.001$ \\
Bible & 35.28 & 1.66 & $p < 0.05$ & NS & $p < 0.05$ & -- & $p < 0.001$ \\
Random & 26.11 & 0.56 & NS & $p < 0.001$ & $p < 0.001$ & $p < 0.001$ & -- \\
\midrule
\multicolumn{8}{c}{\textbf{ESRD}} \\
\midrule
Baseline & 32.22 & 1.70 & -- & $p < 0.05$ & NS & NS & $p < 0.001$ \\
nephSAP & 38.89 & 0.79 & $p < 0.05$ & -- & NS & NS & $p < 0.001$ \\
UpToDate & 36.95 & 1.23 & NS & NS & -- & NS & $p < 0.001$ \\
Bible & 34.17 & 1.15 & NS & NS & NS & -- & $p < 0.001$ \\
Random & 21.95 & 2.00 & $p < 0.001$ & $p < 0.001$ & $p < 0.001$ & $p < 0.001$ & -- \\
\midrule
\multicolumn{8}{c}{\textbf{Electrolyte}} \\
\midrule
Baseline & 25.86 & 4.34 & -- & NS & NS & NS & NS \\
nephSAP & 17.67 & 3.62 & NS & -- & NS & NS & NS \\
UpToDate & 16.81 & 2.48 & NS & NS & -- & NS & NS \\
Bible & 18.10 & 3.19 & NS & NS & NS & -- & NS \\
Random & 23.28 & 0.50 & NS & NS & NS & NS & -- \\
\midrule
\multicolumn{8}{c}{\textbf{Pregnancy}} \\
\midrule
Baseline & 40.00 & 1.92 & -- & NS & $p < 0.05$ & NS & $p < 0.001$ \\
nephSAP & 48.34 & 3.47 & NS & -- & NS & NS & $p < 0.001$ \\
UpToDate & 50.83 & 2.85 & $p < 0.05$ & NS & -- & NS & $p < 0.001$ \\
Bible & 47.50 & 2.10 & NS & NS & NS & -- & $p < 0.001$ \\
Random & 19.17 & 0.83 & $p < 0.001$ & $p < 0.001$ & $p < 0.001$ & $p < 0.001$ & -- \\
\midrule
\multicolumn{8}{c}{\textbf{IR}} \\
\midrule
Baseline & 35.56 & 1.11 & -- & $p < 0.01$ & NS & NS & $p < 0.05$ \\
nephSAP & 48.34 & 0.96 & $p < 0.01$ & -- & NS & NS & $p < 0.001$ \\
UpToDate & 44.17 & 0.84 & NS & NS & -- & NS & $p < 0.001$ \\
Bible & 40.84 & 2.10 & NS & NS & NS & -- & $p < 0.001$ \\
Random & 23.33 & 4.08 & $p < 0.05$ & $p < 0.001$ & $p < 0.001$ & $p < 0.001$ & -- \\
\midrule
\multicolumn{8}{c}{\textbf{Infection}} \\
\midrule
Baseline & 37.78 & 1.11 & -- & NS & NS & NS & $p < 0.001$ \\
nephSAP & 35.84 & 0.84 & NS & -- & NS & NS & $p < 0.001$ \\
UpToDate & 39.17 & 1.59 & NS & NS & -- & NS & $p < 0.001$ \\
Bible & 36.67 & 0.00 & NS & NS & NS & -- & $p < 0.001$ \\
Random & 21.67 & 2.89 & $p < 0.001$ & $p < 0.001$ & $p < 0.001$ & $p < 0.001$ & -- \\
\end{longtable}
\end{center}

\begin{center}
\small
\begin{longtable}{@{}p{1.3cm}ccccccc@{}}
\caption{Llama 3 RAG percent of MCQ answered correctly by subcategory}
\label{tab:llama_results} \\

\toprule
\textbf{Source} & \textbf{Mean (\%)} & \textbf{SEM} & \textbf{vs Baseline} & \textbf{vs nephSAP} & \textbf{vs UpToDate} & \textbf{vs Bible} & \textbf{vs Random} \\
\midrule
\endfirsthead

\multicolumn{8}{c}%
{{\bfseries \tablename\ \thetable{} -- continued from previous page}} \\
\toprule
\textbf{Source} & \textbf{Mean (\%)} & \textbf{SEM} & \textbf{vs Baseline} & \textbf{vs nephSAP} & \textbf{vs UpToDate} & \textbf{vs Bible} & \textbf{vs Random} \\
\midrule
\endhead

\midrule \multicolumn{8}{r}{{Continued on next page}} \\ \midrule
\endfoot

\bottomrule
\endlastfoot

\multicolumn{8}{c}{\textbf{Hypertension}} \\
\midrule
Baseline & 58.43 & 0.00 & -- & NS & NS & NS & $p < 0.001$ \\
nephSAP & 59.83 & 0.96 & NS & -- & NS & NS & $p < 0.001$ \\
UpToDate & 57.87 & 0.73 & NS & NS & -- & NS & $p < 0.001$ \\
Bible & 57.87 & 0.73 & NS & NS & NS & -- & $p < 0.001$ \\
Random & 48.04 & 0.84 & $p < 0.001$ & $p < 0.001$ & $p < 0.001$ & $p < 0.001$ & -- \\
\midrule
\multicolumn{8}{c}{\textbf{Glomerular Diseases}} \\
\midrule
Baseline & 45.31 & 0.43 & -- & $p < 0.001$ & $p < 0.001$ & NS & $p < 0.001$ \\
nephSAP & 54.70 & 0.64 & $p < 0.001$ & -- & $p < 0.002$ & $p < 0.001$ & $p < 0.001$ \\
UpToDate & 50.84 & 0.64 & $p < 0.001$ & $p < 0.002$ & -- & $p < 0.001$ & $p < 0.001$ \\
Bible & 45.47 & 0.50 & NS & $p < 0.001$ & $p < 0.001$ & -- & $p < 0.001$ \\
Random & 33.22 & 0.80 & $p < 0.001$ & $p < 0.001$ & $p < 0.001$ & $p < 0.001$ & -- \\
\midrule
\multicolumn{8}{c}{\textbf{AKI}} \\
\midrule
Baseline & 59.83 & 0.84 & -- & NS & NS & NS & $p < 0.001$ \\
nephSAP & 63.20 & 1.33 & NS & -- & NS & NS & $p < 0.001$ \\
UpToDate & 60.68 & 1.03 & NS & NS & -- & NS & $p < 0.001$ \\
Bible & 59.83 & 1.25 & NS & NS & NS & -- & $p < 0.001$ \\
Random & 44.10 & 0.84 & $p < 0.001$ & $p < 0.001$ & $p < 0.001$ & $p < 0.001$ & -- \\
\midrule
\multicolumn{8}{c}{\textbf{Divalent Ions}} \\
\midrule
Baseline & 48.01 & 0.56 & -- & NS & NS & NS & $p < 0.001$ \\
nephSAP & 49.56 & 1.40 & NS & -- & NS & $p < 0.05$ & $p < 0.001$ \\
UpToDate & 47.57 & 0.22 & NS & NS & -- & NS & $p < 0.001$ \\
Bible & 44.69 & 1.28 & NS & $p < 0.05$ & NS & -- & $p < 0.001$ \\
Random & 33.85 & 0.91 & $p < 0.001$ & $p < 0.001$ & $p < 0.001$ & $p < 0.001$ & -- \\
\midrule
\multicolumn{8}{c}{\textbf{Transplant}} \\
\midrule
Baseline & 51.67 & 1.06 & -- & NS & NS & NS & $p < 0.001$ \\
nephSAP & 52.50 & 0.53 & NS & -- & NS & NS & $p < 0.001$ \\
UpToDate & 50.83 & 1.23 & NS & NS & -- & NS & $p < 0.001$ \\
Bible & 50.28 & 1.23 & NS & NS & NS & -- & $p < 0.001$ \\
Random & 40.00 & 1.20 & $p < 0.001$ & $p < 0.001$ & $p < 0.001$ & $p < 0.001$ & -- \\
\midrule
\multicolumn{8}{c}{\textbf{CKD}} \\
\midrule
Baseline & 63.33 & 1.76 & -- & NS & $p < 0.05$ & NS & $p < 0.001$ \\
nephSAP & 66.39 & 0.83 & NS & -- & NS & $p < 0.05$ & $p < 0.001$ \\
UpToDate & 70.83 & 1.59 & $p < 0.05$ & NS & -- & $p < 0.001$ & $p < 0.001$ \\
Bible & 59.44 & 1.84 & NS & $p < 0.05$ & $p < 0.001$ & -- & $p < 0.01$ \\
Random & 51.67 & 1.32 & $p < 0.001$ & $p < 0.001$ & $p < 0.001$ & $p < 0.01$ & -- \\
\midrule
\multicolumn{8}{c}{\textbf{ESRD}} \\
\midrule
Baseline & 58.33 & 1.32 & -- & NS & NS & NS & $p < 0.001$ \\
nephSAP & 63.06 & 0.95 & NS & -- & NS & $p < 0.01$ & $p < 0.001$ \\
UpToDate & 57.78 & 2.03 & NS & NS & -- & NS & $p < 0.001$ \\
Bible & 54.72 & 0.95 & NS & $p < 0.01$ & NS & -- & $p < 0.001$ \\
Random & 40.56 & 2.05 & $p < 0.001$ & $p < 0.001$ & $p < 0.001$ & $p < 0.001$ & -- \\
\midrule
\multicolumn{8}{c}{\textbf{Electrolyte}} \\
\midrule
Baseline & 47.85 & 1.47 & -- & $p < 0.05$ & NS & NS & $p < 0.001$ \\
nephSAP & 42.24 & 0.50 & $p < 0.05$ & -- & $p < 0.001$ & $p < 0.001$ & $p < 0.001$ \\
UpToDate & 51.29 & 0.83 & NS & $p < 0.001$ & -- & NS & $p < 0.001$ \\
Bible & 50.86 & 1.49 & NS & $p < 0.001$ & NS & -- & $p < 0.001$ \\
Random & 25.00 & 1.11 & $p < 0.001$ & $p < 0.001$ & $p < 0.001$ & $p < 0.001$ & -- \\
\midrule
\multicolumn{8}{c}{\textbf{Pregnancy}} \\
\midrule
Baseline & 65.00 & 0.96 & -- & NS & NS & $p < 0.01$ & NS \\
nephSAP & 61.67 & 2.15 & NS & -- & NS & $p < 0.05$ & NS \\
UpToDate & 55.84 & 0.84 & NS & NS & -- & NS & NS \\
Bible & 52.50 & 3.44 & $p < 0.01$ & $p < 0.05$ & NS & -- & $p < 0.05$ \\
Random & 61.67 & 3.19 & NS & NS & NS & $p < 0.05$ & -- \\
\midrule
\multicolumn{8}{c}{\textbf{IR}} \\
\midrule
Baseline & 49.17 & 2.10 & -- & $p < 0.05$ & NS & $p < 0.05$ & $p < 0.001$ \\
nephSAP & 60.83 & 1.59 & $p < 0.05$ & -- & NS & $p < 0.001$ & $p < 0.001$ \\
UpToDate & 51.67 & 3.97 & NS & NS & -- & $p < 0.05$ & $p < 0.001$ \\
Bible & 38.33 & 2.15 & $p < 0.05$ & $p < 0.001$ & $p < 0.05$ & -- & NS \\
Random & 28.34 & 3.47 & $p < 0.001$ & $p < 0.001$ & $p < 0.001$ & NS & -- \\
\midrule
\multicolumn{8}{c}{\textbf{Infection}} \\
\midrule
Baseline & 53.34 & 1.93 & -- & NS & NS & NS & NS \\
nephSAP & 58.34 & 0.96 & NS & -- & NS & $p < 0.05$ & NS \\
UpToDate & 55.84 & 0.84 & NS & NS & -- & NS & NS \\
Bible & 51.67 & 0.96 & NS & $p < 0.05$ & NS & -- & NS \\
Random & 55.00 & 1.67 & NS & NS & NS & NS & -- \\
\end{longtable}
\end{center}

\begin{center}
\small
\begin{longtable}{@{}p{1.3cm}ccccccc@{}}
\caption{Phi-3 RAG percent of MCQ answered correctly by subcategory}
\label{tab:phi3_results} \\

\toprule
\textbf{Source} & \textbf{Mean (\%)} & \textbf{SEM} & \textbf{vs Baseline} & \textbf{vs nephSAP} & \textbf{vs UpToDate} & \textbf{vs Bible} & \textbf{vs Random} \\
\midrule
\endfirsthead

\multicolumn{8}{c}%
{{\bfseries \tablename\ \thetable{} -- continued from previous page}} \\
\toprule
\textbf{Source} & \textbf{Mean (\%)} & \textbf{SEM} & \textbf{vs Baseline} & \textbf{vs nephSAP} & \textbf{vs UpToDate} & \textbf{vs Bible} & \textbf{vs Random} \\
\midrule
\endhead

\midrule \multicolumn{8}{r}{{Continued on next page}} \\ \midrule
\endfoot

\bottomrule
\endlastfoot

\multicolumn{8}{c}{\textbf{Hypertension}} \\
\midrule
Baseline & 60.40 & 0.54 & -- & $p < 0.05$ & $p < 0.003$ & NS & $p < 0.001$ \\
nephSAP & 55.90 & 0.71 & $p < 0.05$ & -- & NS & NS & NS \\
UpToDate & 54.20 & 1.48 & $p < 0.003$ & NS & -- & $p < 0.05$ & NS \\
Bible & 59.00 & 0.56 & NS & NS & $p < 0.05$ & -- & $p < 0.004$ \\
Random & 53.10 & 1.48 & $p < 0.001$ & NS & NS & $p < 0.004$ & -- \\
\midrule
\multicolumn{8}{c}{\textbf{Glomerular Diseases}} \\
\midrule
Baseline & 48.50 & 1.53 & -- & NS & NS & NS & $p < 0.001$ \\
nephSAP & 50.30 & 0.55 & NS & -- & $p < 0.05$ & $p < 0.05$ & $p < 0.001$ \\
UpToDate & 44.50 & 0.57 & NS & $p < 0.05$ & -- & NS & NS \\
Bible & 44.50 & 1.21 & NS & $p < 0.05$ & NS & -- & NS \\
Random & 40.30 & 1.67 & $p < 0.001$ & $p < 0.001$ & NS & NS & -- \\
\midrule
\multicolumn{8}{c}{\textbf{AKI}} \\
\midrule
Baseline & 57.60 & 0.84 & -- & NS & $p < 0.001$ & $p < 0.001$ & $p < 0.001$ \\
nephSAP & 54.80 & 0.71 & NS & -- & $p < 0.05$ & $p < 0.05$ & $p < 0.001$ \\
UpToDate & 51.40 & 1.06 & $p < 0.001$ & $p < 0.05$ & -- & NS & $p < 0.001$ \\
Bible & 51.10 & 0.73 & $p < 0.001$ & $p < 0.05$ & NS & -- & $p < 0.001$ \\
Random & 45.20 & 0.54 & $p < 0.001$ & $p < 0.001$ & $p < 0.001$ & $p < 0.001$ & -- \\
\midrule
\multicolumn{8}{c}{\textbf{Divalent Ions}} \\
\midrule
Baseline & 44.50 & 2.61 & -- & NS & NS & NS & $p < 0.05$ \\
nephSAP & 48.50 & 0.98 & NS & -- & NS & NS & $p < 0.002$ \\
UpToDate & 45.60 & 0.85 & NS & NS & -- & NS & $p < 0.05$ \\
Bible & 43.40 & 2.01 & NS & NS & NS & -- & NS \\
Random & 37.60 & 1.79 & $p < 0.05$ & $p < 0.002$ & $p < 0.05$ & NS & -- \\
\midrule
\multicolumn{8}{c}{\textbf{Transplant}} \\
\midrule
Baseline & 46.90 & 0.95 & -- & NS & NS & NS & NS \\
nephSAP & 46.70 & 1.20 & NS & -- & NS & NS & NS \\
UpToDate & 46.10 & 1.73 & NS & NS & -- & NS & NS \\
Bible & 49.40 & 0.96 & NS & NS & NS & -- & $p < 0.005$ \\
Random & 42.80 & 0.96 & NS & NS & NS & $p < 0.005$ & -- \\
\midrule
\multicolumn{8}{c}{\textbf{CKD}} \\
\midrule
Baseline & 55.80 & 1.15 & -- & NS & NS & $p < 0.05$ & $p < 0.001$ \\
nephSAP & 54.40 & 0.79 & NS & -- & NS & NS & $p < 0.001$ \\
UpToDate & 52.20 & 0.91 & NS & NS & -- & NS & $p < 0.001$ \\
Bible & 49.70 & 2.00 & $p < 0.05$ & NS & NS & -- & $p < 0.005$ \\
Random & 41.70 & 1.95 & $p < 0.001$ & $p < 0.001$ & $p < 0.001$ & $p < 0.005$ & -- \\
\midrule
\multicolumn{8}{c}{\textbf{ESRD}} \\
\midrule
Baseline & 48.10 & 1.66 & -- & NS & NS & NS & $p < 0.001$ \\
nephSAP & 46.40 & 1.23 & NS & -- & NS & NS & $p < 0.01$ \\
UpToDate & 43.90 & 0.96 & NS & NS & -- & NS & NS \\
Bible & 46.10 & 1.16 & NS & NS & NS & -- & $p < 0.05$ \\
Random & 39.40 & 1.60 & $p < 0.001$ & $p < 0.01$ & NS & $p < 0.05$ & -- \\
\midrule
\multicolumn{8}{c}{\textbf{Electrolyte}} \\
\midrule
Baseline & 45.70 & 1.11 & -- & NS & $p < 0.05$ & $p < 0.005$ & $p < 0.05$ \\
nephSAP & 39.20 & 3.26 & NS & -- & NS & NS & NS \\
UpToDate & 37.90 & 1.22 & $p < 0.05$ & NS & -- & NS & NS \\
Bible & 34.90 & 1.47 & $p < 0.005$ & NS & NS & -- & NS \\
Random & 37.50 & 1.91 & $p < 0.05$ & NS & NS & NS & -- \\
\midrule
\multicolumn{8}{c}{\textbf{Pregnancy}} \\
\midrule
Baseline & 67.50 & 2.85 & -- & NS & NS & NS & $p < 0.004$ \\
nephSAP & 56.70 & 2.36 & NS & -- & NS & NS & NS \\
UpToDate & 69.20 & 2.50 & NS & NS & -- & NS & $p < 0.002$ \\
Bible & 60.00 & 4.08 & NS & NS & NS & -- & NS \\
Random & 48.30 & 4.41 & $p < 0.004$ & NS & $p < 0.002$ & NS & -- \\
\midrule
\multicolumn{8}{c}{\textbf{IR}} \\
\midrule
Baseline & 51.70 & 2.15 & -- & $p < 0.01$ & $p < 0.003$ & NS & NS \\
nephSAP & 64.20 & 2.85 & $p < 0.01$ & -- & NS & $p < 0.001$ & $p < 0.001$ \\
UpToDate & 65.80 & 2.10 & $p < 0.003$ & NS & -- & $p < 0.001$ & $p < 0.001$ \\
Bible & 47.50 & 3.44 & NS & $p < 0.001$ & $p < 0.001$ & -- & NS \\
Random & 47.50 & 0.83 & NS & $p < 0.001$ & $p < 0.001$ & NS & -- \\
\midrule
\multicolumn{8}{c}{\textbf{Infection}} \\
\midrule
Baseline & 50.80 & 0.83 & -- & $p < 0.05$ & $p < 0.05$ & NS & $p < 0.004$ \\
nephSAP & 59.20 & 2.10 & $p < 0.05$ & -- & $p < 0.001$ & NS & $p < 0.001$ \\
UpToDate & 43.30 & 1.36 & $p < 0.05$ & $p < 0.001$ & -- & $p < 0.001$ & NS \\
Bible & 57.50 & 1.60 & NS & NS & $p < 0.001$ & -- & $p < 0.001$ \\
Random & 40.00 & 3.04 & $p < 0.004$ & $p < 0.001$ & NS & $p < 0.001$ & -- \\
\end{longtable}
\end{center}
\end{document}